\documentclass[10pt,twocolumn,letterpaper]{article}

\usepackage[pagenumbers]{cvpr} 

\usepackage{float}
\usepackage{siunitx}
\usepackage{stfloats}
\usepackage[dvipsnames]{xcolor}
\usepackage[percent]{overpic}

\definecolor{BBColor}{rgb}{0.0,0.4,0.4}

\definecolor{RanColor}{rgb}{0,0,1}

\newcommand{\Ss}{\mathcal{S}^2}
\newcommand{\win}{\omega_\mathrm{i}}
\newcommand{\wout}{\omega_\mathrm{o}}
\newcommand{\Le}{L_\mathrm{e}}
\newcommand{\Ld}{L_\mathrm{dir}}
\newcommand{\Li}{L_\mathrm{ind}}
\newcommand{\Td}{\mathcal{T}_\mathrm{dir}}
\newcommand{\Ti}{\mathcal{T}_\mathrm{ind}}

\newif\ifhascapturedata
\hascapturedatafalse 
\newcommand{\hascaptureversion}[1]{\ifhascapturedata #1\fi}

\definecolor{cvprblue}{rgb}{0.21,0.49,0.74}
\usepackage[pagebackref,breaklinks,colorlinks,citecolor=cvprblue]{hyperref}

\title{BiGS: Bidirectional Gaussian Primitives for Relightable 3D Gaussian Splatting}

\author{Zhenyuan Liu\textsuperscript{1,2}, Yu Guo\textsuperscript{3}, Xinyuan Li\textsuperscript{1}, Bernd Bickel\textsuperscript{2}, Ran Zhang\textsuperscript{1}\\\\
\textsuperscript{1}Tencent PCG, New York, USA\\
\textsuperscript{2}ETH Zurich, Zurich, Switzerland\\
\textsuperscript{3}George Mason University, Fairfax, USA\\\\
{\tt\small zhenyliu@ethz.ch, tflsguoyu@gmail.com, xinyuanli@global.tencent.com,}\\
{\tt\small bickelb@ethz.ch, ranorizhang@global.tencent.com}
\and
}

\begin{document}
\twocolumn[{
    \renewcommand\twocolumn[1][]{#1}
    \maketitle
    \begin{center}
    \centering
    \captionsetup{type=figure}
    \includegraphics[width=\linewidth]{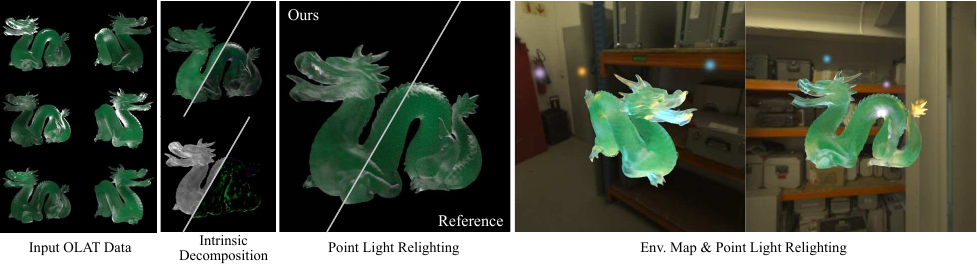}
    \vspace{-1em}
    \captionof{figure}{BiGS reconstructs view- and light-dependent color functions for Gaussian Splats using OLAT data such as the shown translucent \textsc{Dragon} of 31,252 bidirectional Gaussian primitives. Our model decomposes the appearance of each primitive into different intrinsic components and is able to achieve plausible relighting and novel view synthesis with various environment maps and point light sources.}
    \label{fig:teaser}
\end{center}
}]
\begin{abstract}
We present Bidirectional Gaussian Primitives,
an image-based novel view synthesis technique designed to represent and render 3D objects with surface and volumetric materials under dynamic illumination.
Our approach integrates light intrinsic decomposition into the Gaussian splatting framework, enabling real-time relighting of 3D objects.
To unify surface and volumetric material within a cohesive appearance model, we adopt a light- and view-dependent scattering representation via bidirectional spherical harmonics.
Our model does not use a specific surface normal-related reflectance function, making it more compatible with volumetric representations like Gaussian splatting, where the normals are undefined.
We demonstrate our method by reconstructing and rendering objects with complex materials. Using One-Light-At-a-Time (OLAT) data as input, we can reproduce photorealistic appearances under novel lighting conditions in real time.
\end{abstract}    
\section{Introduction}
\label{sec:intro}
Capturing and rendering 3D content from images has been a long-standing research topic in computer graphics and computer vision, with a wide range of applications in virtual production, video games, architecture, and mixed reality.

Recently, 3D Gaussian Splatting (3DGS) \cite{kerbl_3d_2023} has emerged as a novel 3D representation that excels in these areas, delivering both a high level of photo-realism and real-time performance.
By representing scenes with a set of anisotropic Gaussian primitives, it can effectively capture complex visual details and render them quickly. 
However, the spherical harmonics based appearance model used in 3DGS is only capable of synthesizing novel views under the static illumination in the capture data, unable to adapt to new lighting conditions.
This limits the usage of 3DGS in interactive applications like video games and mixed reality, where dynamic lighting plays an essential role.

Recent research \cite{jiang_gaussianshader_2023, gao_relightable_2023} has attempted to address this limitation by incorporating surface-based shading models into the framework of Gaussian Splatting, describing how light interacts with Gaussian primitives.
These models can work well for objects with surface-based materials,
such as smooth or rough solid objects. 
However, they tend to struggle with fuzzy objects that lack a clear surface definition, like fur and hair, or those with volumetric appearances that are not solely determined by their surface.
For example, subsurface scattering in translucent objects presents significant challenges for surface-based models.

In this paper, we introduce a lighting-dependent appearance model for Gaussian primitives based on bidirectional spherical harmonics. As shown in~\cref{fig:teaser}, our approach enables the rendering of 3D objects with various materials under dynamic illumination, including both near-field and environmental lighting.
Our unified representation does not make assumption of the materials of the objects, therefore enabling the modeling of both surface-based and volumetric appearance.
By modeling the light model as discussed in~\cref{sec:background-gs},
we compute the diffused scattering and directional scattering components as an intrinsic decomposition of lighting-dependent radiance in~\cref{ssec:light-decomposition}.
With a bidirectional spherical harmonics representation, described in~\cref{ssec:sh-representation},
we efficiently model and render complex materials under dynamic lighting conditions, as detailed in~\cref{ssec:rendering}.
Finally, in~\cref{sec:optimization}, we present our physics-inspired regularization terms that lead to a more stable optimization process of our model on our OLAT data, and experiments with synthetic data and light-stage capture data as shown in~\cref{sec:results}.
To summarize, our contributions include:
\begin{itemize}
    \item A novel formulation of relightable Gaussian primitives, accounting for both surface and volume appearances.
    \item Bidirectional spherical harmonics for representing the light--dependent scattering function for Gaussian primitives.
    \item An optimization method for obtaining relightable Gaussian primitives from OLAT datasets.
\end{itemize}

\section{Related Work}
\label{sec:related-work}

\subsection{Relightable Gaussian Splatting}
Recently, Gaussian Splatting has become one of the most popular techniques in novel view synthesis.
It represents scenes using 3D Gaussian volumetric primitives and reaches high-level photorealism and real-time rendering performance via fast rasterization.
One major limitation of Gaussian Splatting is that the illumination in the training data is baked into the model, 
making it challenging to render the scene under novel lighting conditions.
Saito et al. \cite{saito_relightable_2023} decomposes the color of the Gaussians into various components using multi-layer perceptrons (MLPs) for modeling human faces.
SuGar \cite{guedon_sugar_2023} and 2D Gaussian Splatting \cite{huang_2d_2024} assume fully opaque surfaces and extract meshes and textures from Gaussian Splats by optimizing the shape of the Gaussian primitives from ellipsoids to planar disks. 
Gao et al. \cite{gao_relightable_2023} proposes to derive a normal map from a depth field, and then extract shading properties using physics-based rendering.  
GaussianShader \cite{jiang_gaussianshader_2023} estimates the normals of the Guassians using their shapes and proposes a shading model for reflective surfaces. 
GS-Phong \cite{he_gs-phong_2024} also assumes fully opaque Gaussians and applies the Blinn-Phong shading model to compute diffuse and specular colors.
These methods all adopt surface-based appearance models, while our method sticks with a general lighting formulation that can represent both surface and volumetric effects, hence without the need to drive all the Gaussians to be completely opaque or rely on a specific shading model.
While there are also some works \cite{condor_volumetric_2024, zhou_unified_2024} using Gaussian as primitives for volumetric ray tracing, reaching outstanding photorealism at a greater computation cost, our method is rasterization-based, without compromising real-time rendering speed.

\subsection{Relighting of Neural Implicit Representation}
Neural radiance field (NeRF) \cite{mildenhall_nerf_2020, barron_mip-nerf_2022} is another popular representation for novel view synthesis. It uses MLP to represent spatially varying color and density fields, and apply ray marching to render. 
NeRF faces similar limitations as Gaussian Splatting: The illumination, geometry, and material of the objects in the training data are baked into the radiance fields. 
To separate the illumination and material, many researchers try to incorporate various material and lighting representations into NeRF by conditioning the color field on intermediate quantities such as visibility and lighting \cite{xu_renerf_2023, toschi_relight_2023} with a material/lighting model, 
leading to more efficient rendering and relighting of objects of complex appearance such as reflective surfaces \cite{srinivasan_nerv_2020, zhang_nerfactor_2021, zheng_neural_2021, verbin_ref-nerf_2021}, self-emissive \cite{jeong_esr-nerf_2024} and transparent objects \cite{wang_nemto_2023}.
Surface representations such as neural signed distance function \cite{wang_neus_2021, yariv_volume_2021} are also introduced to disentangle geometry from appearance, enabling challenging material modeling and high-quality surface reconstruction simultaneously \cite{yariv_bakedsdf_2023, liu_nero_2023, ge_ref-neus_2023, cai_neuralto_2024}. Using light transport hints generated from signed distance functions, Zeng et al. \cite{zeng_relighting_2023} also demonstrates relighting NeRF with high-frequency shadows and highlights.
Other than surface materials, a range of scattering models for volumes are also incorporated to relight translucent objects \cite{yu_learning_2023, zhu_neural_2023}. Zhang et al. \cite{zhang_nemf_2023} uses the SGGX phase function \cite{heitz_sggx_2015} to achieve a parameterized subsurface scattering appearance.
LitNeRF \cite{sarkar_litnerf_2023} decomposes the lighting into reflectance and intrinsic components and demonstrates physically plausible relighting performance for human faces. Our method shares the same spirit with LitNeRF in terms of the lighting model, but we only use spherical harmonics for representing different components in our model instead of MLP, and we also introduce additional regularization terms to disambiguate multiple solutions in the decomposition.
A recent study by Lyu et al. \cite{lyu_neural_2022} borrows the idea of precomputed radiance transfer \cite{sloan_precomputed_2002, ng_triple_2004} to enhance the relighting capability of NeRF considering global illumination.




\section{Background: 3D Gaussian Splatting}
\label{sec:background-gs}

\begin{figure*}[ht]
    \centering
    \begin{overpic}[width=\linewidth]{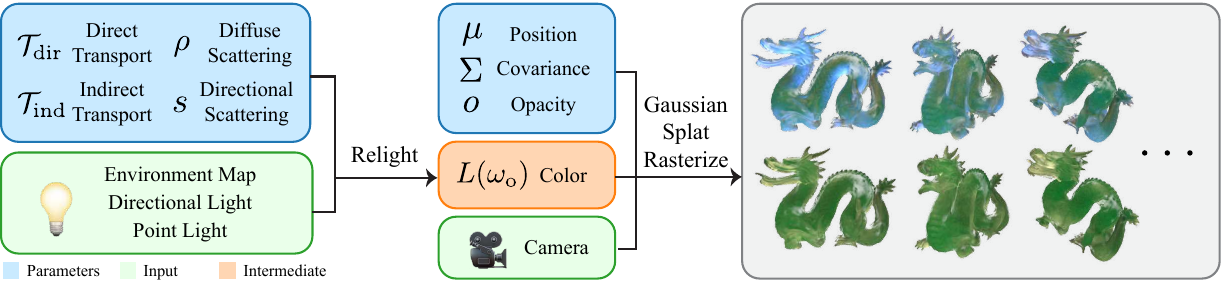}
    \put(28.3, 6.7){\cref{eq:out-radiance-sum}}
    \put(53.5, 6.7){\cref{eq:gs}}
    \end{overpic}
    \caption{\textbf{Pipeline overview}: our method introduces per-Gaussian optimizable lighting parameters: $\Td$, $\Ti$, $\rho$, and $s$, each represented using spherical harmonics. Given novel lighting conditions, we relight each Gaussian by generating the view-dependent color of each Gaussian $L(\wout)$ represented by spherical harmonics that are compatible with the Gaussian rasterization pipeline, and therefore can render under novel light and view conditions.}
    \label{fig:pipeline-overview}
\end{figure*}

3D Gaussian Splatting (3DGS) \cite{kerbl_3d_2023} represents a scene using numerous volumetric primitives in the form of 3D Gaussian kernels.  
Each Gaussian primitive is characterized by a set of parameters, including its position $\mu$ (center of the primitive), covariance matrix $\Sigma$ (parameterized by scale and rotation), opacity $o$ at the center, and spherical harmonics coefficients that define the color function $L(\wout)$, which depends on the view direction $\wout$.

To render a scene represented by $n$ 3D Gaussian primitives, all primitives are projected onto the camera plane, resulting in 2D Gaussian kernels with mean $\mu'_i$ and covariance $\Sigma'_i$.  
The color of a pixel centered at $p$ is then determined by $\alpha$-blending the projected Gaussians from the nearest to the farthest relative to the camera, following the equation:
\begin{equation}
    I(\wout) = \sum_{i = 1}^{n} T_i \alpha_i L_{i} (\wout), \; T_i = \prod_{j = 1}^{i - 1} (1 - \alpha_j).
    \label{eq:gs}
\end{equation}
Here, $\alpha_i = o_i G_i(p)$ accounts for the falloff induced by the 2D Gaussian, where $G_i(p) = \exp \left(-\frac{1}{2} (p - \mu_i')^t \Sigma_i'^{-1} (p - \mu_i')\right)$.  
The term $T_i$ denotes the transmittance of from the $i$-th Gaussian primitive to the camera, while the color function $L_i(\wout)$ encodes view-dependent color.
This color function is represented by spherical harmonics.  
However, in ~\cref{eq:gs}, the color function is solely dependent on the view direction, making it only suitable for reconstructing and rendering objects under static illumination.


To achieve the goal of reproducing appearance under dynamic illumination,
We extend the color function in \cref{eq:gs}
 to model lighting-dependent Gaussian primitives.
As illustrated in \cref{fig:pipeline-overview} and detailed in the following sections, the extended color function $L(\wout)$ incorporates additional parameters, including direct and indirect light transport, as well as diffuse and directional scattering components.
This allows us to take the input of diverse lighting conditions—including environment maps, directional lights, and point lights—along with camera settings, to render Gaussian splats under dynamic illumination. 






\section{Relightable Gaussian Primitives}
\label{sec:method}

\subsection{Intrinsic Light Decomposition}
\label{ssec:light-decomposition}
The key to relightable Gaussian primitives is to express the view-dependent color $L(\wout)$ of each Gaussian as a function of both view direction and the lighting conditions,
including the direction from the light source to the Gaussian $\win$ and the light source's intensity in $\win$, denoted by $\Le(\win)$. 
This approach allows us to update $L(\wout)$ when the lighting changes, and therefore to render under novel lighting conditions.

Inspired by LitNeRF \cite{sarkar_litnerf_2023}, we adopt the intrinsic decomposition model that partitions the color of each Gaussian into the sum of two components, assuming no self-emission:
\begin{equation}
    L(\wout) = \Ld(\wout) + \Li,
\end{equation}
where $\Ld$ stands for the direct illumination -- the contribution from the light that travels from the emitter to the Gaussian directly and $\Li$ for the indirect illumination that encompasses the light bounces and scatters in the scene and arrives at the Gaussian. We follow the common assumption that $\Li$ does not depend on view direction \cite{sarkar_litnerf_2023}.

We model the direct illumination by modeling the attenuation of light from the emitter that scatters into view direction,
\begin{equation}
    \Ld(\wout) = \int_{\Ss} \Td(\win) \Le(\win) f(\win, \wout) \;\mathrm{d} \win,
    \label{eq:direct-light-with-phase}
\end{equation}
 where $\Td(\win): \Ss \rightarrow \mathbb{R}$ is the direct transport operator, which models the ratio of the light arriving from the emitter to the Gaussian in $\win$, accounting for attenuation from occlusion or absorption along the light direction; $\Le(\win)$ is the light intensity at the emitter. 
The product of $\Td(\win)$ and $\Le (\win)$ essentially represents the amount of direct radiance the Gaussian receives from direction $\win$.
$f(\win, \wout): \Ss \times \Ss \rightarrow \mathbb{R}^3$ is the scattering function, indicating the amount of scattering per RGB channel when light travels in from $\win$ and out into $\wout$. 
Deviating from LitNeRF \cite{sarkar_litnerf_2023}, we choose to use 3 channels for the output of $f$ to capture the colorful highlight that can be observed in metallic or iridescent reflection, as shown in \cref{fig:phase-channels}.

For scattering functions to be physically meaningful, they have to satisfy properties such as being reciprocal,
\begin{equation}
    f(\win, \wout) = f(\wout, \win) \;\; \forall \win, \wout.
    \label{eq:phase-reciprocity}
\end{equation}

Reciprocity results from the reversibility of light direction. Another property of scattering functions
originates from energy conservation -- the amount of light arriving at a Gaussian cannot exceed the total amount of light leaving this Gaussian, leading to $f$ of norm less than 1,
\begin{equation}
    \int_{\Ss} f(\win, \wout) \; \mathrm{d}\wout \leq 1 \;\; \forall \win.
    \label{eq:phase-unit-norm}
\end{equation}

In BiGS, we explicitly enforce reciprocity  to reduce the number of optimizable parameters and convert energy conservation into a penalty term to achieve a more stable decomposition and optimization,
as will be discussed \cref{ssec:model-supervision}. 
We further decompose $f$ into two parts: a diffuse component $\rho \in \mathbb{R}^3$, which captures lighting- and view-independent features such as albedo and ambient occlusion, and a directional scattering component $s(\win, \wout)$, which models the lighting- and view-dependent aspects.
With this decomposition, we have \cref{eq:direct-light-with-phase} rewritten as the following for direction illumination:
\begin{equation}
    \Ld(\wout) = \int_{\Ss} \Td(\win) \Le(\win) (\rho + s(\win, \wout)) \;\mathrm{d} \win.
    \label{eq:direct-illumination}
\end{equation}

As for indirect illumination $\Li$, we model it as a residue term that captures the direct illumination struggles to represent. We express it as 
\begin{equation}
    \Li = \int_{\Ss} \Ti(\win) \Le(\win) \;\mathrm{d} \win,
    \label{eq:indirect-illumination}
\end{equation}
where $\Ti: \Ss \rightarrow \mathbb{R}^3 $ is the indirect transport operator that models the light travels from the emitter to the Gaussian after multiple bounces.
This term uses RGB channels, unlike $\Td$ to account for the color changes that happen during bounces. 

Different components in the intrinsic decomposition might lead to multiple solutions that produce similar final rendering, and we discuss how we regularize the components and disambiguate the solutions in \cref{ssec:model-supervision}. 

\subsection{Bidirectional Spherical Harmonics}
\label{ssec:sh-representation}

In \cref{ssec:light-decomposition}, we prepare our formulation of decomposing the light into multiple components $\Td(\cdot), s(\cdot, \cdot), \Ti(\cdot)$ and $\rho$. Many of these quantities are defined over $\Ss$ or $\Ss \times \Ss$. Next, we show how we use spherical harmonics to represent $\Td(\cdot)$ and $\Ti(\cdot)$, and spherical harmonics extended to inputs of 2 directions, dubbed \textit{bidirectional spherical harmonics} to represent $s(\cdot, \cdot)$. Then we show how to use our representations to compute Gaussians' colors that can be used for rasterization from arbitrary viewpoints. 

Spherical harmonics are a special set of orthonormal basis functions defined over the surface of a unit sphere in 3D, serving as a great tool in real-time rendering \cite{sloan_precomputed_2002, lehtinen_framework_2007}.
Using the first $n$ spherical harmonics basis functions $\{ y_i(\cdot) \}_{i = 1}^{n}$, a function defined over $\Ss$ can be represented by the sum of the bases and the $n$ corresponding coefficients.
Using $\Td(\cdot)$ as an example: given a direction $\omega$, $\Td(\omega)$ can be evaluated as follows:
\begin{equation}
    \Td(\omega) = \sum_{i = 1}^{n} c_i y_i (\omega),
    \label{eq:sh-reconstruction}
\end{equation}
where $c_i$ is the coefficient corresponding to $y_i$. Likewise, we can represent $\Ti(\cdot)$ in the same fashion.

For each Gaussian, we use the first 25 spherical harmonics basis functions to model $\Td(\cdot)$ and each channel of $\Ti(\cdot)$. This amounts to 25 coefficients for $\Td(\cdot)$ and 75 for $\Ti(\cdot)$ (25 per RGB channel). These coefficients are the optimizable parameters in the training pipeline.

Now we hope to find a representation for functions defined over $\Ss \times \Ss$, for instance, $s(\cdot, \cdot)$. We can do this by composing two groups of spherical harmonics to form a new basis function for $\Ss \times \Ss$.  Using the first $n$ spherical harmonics basis, $s$ can be written as
\begin{equation}
    s(\win, \wout) = \sum_{i = 1}^{n} \sum_{j = 1}^{n} c_{ij} y_i(\win) y_j(\wout),
    \label{eq:double-sh}
\end{equation}
and we call $y_i(\win) y_j(\wout)$ \textit{bidirectional spherical harmonics}. This leads to $n^2$ coefficients $\{ c_{ij} : \forall i, j = 1, 2, \cdots, n \}$. But $n^2$ would have been more than we actually need.
\cref{eq:double-sh} does not enforce the reciprocity of $s$ (\cref{eq:phase-reciprocity}). We incorporate this important physical property into $s$
by letting $c_{ij} = c_{ji}$ for all $i, j$ pairs. This reduces the space of function that can be represented from $\Ss \times \Ss$ to those that are reciprocal; see \cref{sec:reciprocity} in supplementary material for a proof.
And it also makes our model more compact by reducing the number of parameters from $n^2$ to $n (n + 1) / 2$. Using 25 bases, $s$ costs $325 \times 3 = 975$ parameters per primitive.

One could evaluate the amount of scattering $s(\win, \wout)$ given $\win$ and $\wout$, and this results in evaluating $s$ every time viewpoint ($\wout$) changes.
The spherical harmonics representation also allows us to evaluate $s$ over one input variable $\win$ for relighting and later evaluate over $\wout$ for novel view synthesis, so that we only need the second part of evaluation if only aim to render from a new view without changing the light.
Specifically, given light entering the Gaussian from direction $\win$, we can express the out-scatter color as a function over all $\wout$, denoted by $s_{\win}(\wout) : \Ss \rightarrow \mathbb{R}^3$. 
Again, $s_{\win}$ can be represented by spherical harmonics: 
\begin{equation}
    s_{\win} (\cdot) = \sum_{j=0}^n c_j y_j(\cdot),
    \label{eq:sh-partial-eval}
\end{equation}
where coefficients $c_j = \sum_{i=1}^{n} c_{ij} y_{i}(\win)$ come from summing up bases for $\win$. 

\subsection{Rendering Under Novel Lighting Conditions}
\label{ssec:rendering}

Given multiple directional lights in the direction of $\win{}_0, \win{}_1, \cdots,$ $\win{}_t, \cdots$ and their corresponding light intensity $L_e(\win{}_t)$. We evaluate \cref{eq:direct-illumination} and \cref{eq:indirect-illumination} for direct and indirect illumination respectively, where the integral over $\Ss$ becomes the summation of contribution from each light in the scene, each characterized by its direction $\win$ and intensity $L_e(\win)$.
The final color becomes the contribution from each light source:
\begin{equation}
    L(\wout) = \sum_{\win{}_t} \left[ \Td(\win{}_t) (\rho + s_{\win{}_t}(\wout)) + \Ti(\win{}_t)) L_e(\win{}_t) \right] .
    \label{eq:out-radiance-sum}
\end{equation}
Here we evaluate $\Td$ and $\Ti$ by \cref{eq:sh-reconstruction}, 
and partially evaluate $s(\win{}_t, \wout)$ to $s_{\win{}_t}(\wout)$ as per \cref{eq:sh-partial-eval}.
This enables our model to generate spherical harmonics coefficients that are compatible with the original Gaussian Splatting rasterization formulation. 
After we compute $L$ per Gaussian, to render the relighted image $I$, it follows \cref{eq:gs} in \cref{sec:background-gs} to compute the color of each pixel given camera parameters.
Therefore, we compute \cref{eq:out-radiance-sum} \textit{only if} light conditions change. If there is a change only in view direction (such as a new camera position or pose), we just need to render the scene again using the Gaussian Splatting rasterization pipeline, without the need to re-invoke our relighting pipeline. 

The above discussion works with directional lights, and we can also extend it to point lights and environment map light sources. 
To relight with a point light source centered at $\mathbf{x}_t$ and center light intensity $\hat{L}_t$, substitute into \cref{eq:out-radiance-sum}  $\;\win{}_t = (\mathbf{x}_t - \mathbf{\mu}) / || \mathbf{x}_t - \mathbf{\mu} ||$ and, $L_e(\win{}_t) = \hat{L}_t / || \mathbf{x}_t - \mathbf{\mu} ||^2$ where $\mathbf{\mu}$ is the Gaussian center position and $|| \:\: \cdot \:\: ||$ denotes Euclidean norm.

To relight with an environment map, we employ a low-frequency approximation by sampling the environment map at a set of predefined lighting positions $\mathbf{x}_t$. The corresponding light direction is given by $\win{}_t = (\mathbf{x}_t - \mathbf{\mu}) / ||\mathbf{x}_t - \mathbf{\mu}||$, where the light intensity $\Le(\win{}_t)$ is determined by the pixel value of the environment map, weighted by the solid angle associated with $\win{}_t$.

\section{Training BiGS on OLAT data}
\label{sec:optimization}
To construct relightable Gaussian primitives, we employ an inverse optimization method. Given a set of images\hascaptureversion{--whether from real captures or virtual renderings--} in the format of One-Light-At-a-Time (OLAT) data as shown in \cref{fig:teaser}~(Input OLAT Data), we search for the optimal configuration of each Gaussian primitive to best match the provided input images.
\subsection{Model Supervision and Disambiguation}
\label{ssec:model-supervision}
Our method introduces the following optimizable parameters for each Gaussian primitive: $\Td$, $\rho$, $s$, and $\Ti$. These parameters are optimized alongside the original Gaussian Splatting parameters, including position $\mu$, covariance $\Sigma$, and opacity $o$ for each Gaussian.

We supervise our model with two goals in mind: realistic image synthesis under novel lighting conditions, 
and physically plausible intrinsic light decomposition. 
These two goals lead to two terms in our loss function: 
image reconstruction loss $\mathcal{L}_\text{rec}$ and regularization loss $\mathcal{L}_\text{reg}$ of the intrinsic light decomposition
\begin{equation}
\mathcal{L} = \mathcal{L}_\text{rec} + \mathcal{L}_\text{reg},
\end{equation}
For the reconstruction loss, we use the same term in Gaussian Splatting~\cite{kerbl_3d_2023}, 
$\mathcal{L}_\text{rec} = \mathcal{L}_1 + \lambda_{\text{D-SSIM}}\mathcal{L}_\text{D-SSIM}$,
$\mathcal{L}_1, \mathcal{L}_\text{D-SSIM}$ are respectively the mean absolute error and SSIM loss, evaluated on the rendered $\Gamma(I)$ and the reference image $\Gamma(I')$ after applying the tone-mapping function $\Gamma$, as we are using HDR images for training our model. $\lambda_{\text{D-SSIM}}$ is the weight for $\mathcal{L}_\text{D-SSIM}$.

However, only $\mathcal{L}_\text{rec}$ does not guarantee that we obtain plausible outputs for each component in intrinsic light decomposition.
Optimizing \cref{eq:direct-illumination} without regularization might lead to unphotorealistic ambiguity on $\Td(\win)$ and $s(\win, \wout)$ and unstable training.
Therefore, we add the following regularization terms that tackle the energy conservation of the scattering function and the non-negativity of light intensity:
\begin{equation}
\mathcal{L}_\text{reg} =\lambda_\mathrm{s} \mathcal{L}_\mathrm{s} + \lambda_{+} \mathcal{L}_{+}
\end{equation}

We alleviate the ambiguity issue by using the energy conservation constraint in~\cref{eq:phase-unit-norm} as a penalty term.
\begin{equation}
    \mathcal{L}_\mathrm{s} = \frac{1}{N} \left(\int_{\Ss} s_{\win}(\wout) \; \mathrm{d} \wout - 1 \right)_{+}^2,
\label{eq:norm-constraint}
\end{equation}
where $( \cdot )_+ = \max \{ \cdot , 0 \}$
refers to the ReLU function, and $N$ is the number of Gaussians.

During training, $\mathcal{L}_\mathrm{s}$ is evaluated twice with different $\win, \wout$: first by using the training OLAT light direction $\win$ and camera direction $\wout$ associated with the training image; To help generalize the constraint to novel lights and views unseen during training, we also randomly sample a pair of $\win, \wout$ to evaluate the loss with, then add up with the first evaluation. 

The second regularization we consider here is the non-negativity of light intensity. 
We explicitly constrain $\rho$ between 0 and 1 by using a Sigmoid function.
For other lighting components, namely $s, \Td$ and $\Ti$,
we clamp the negative part after computing their values by $\max \{\cdot, 0\}$ before rendering the image.
However, the clamping operation causes the gradients of the negative values to be zero, and therefore they do not get updated during the training loop. We use another loss term $\mathcal{L}_+$ to encourage the values to be non-negative:
\begin{equation}
    \mathcal{L}_{+} = \frac{1}{N} \left[ \Td(\win)_{-}^2 + \Ti(\win)_{-}^2 + s(\win, \wout)_{-}^2 \right],
\label{eq:nonneg-constraint}
\end{equation}
where $( \cdot )_{-} = \min \{ \cdot , 0 \}$. This term is evaluated in the same manner as $\mathcal{L}_s$: on both training $\win, \wout$ and a pair of randomly sampled $\win, \wout$ per iteration.

\begin{figure*}[ht]
    \centering
    \includegraphics[width=\linewidth]{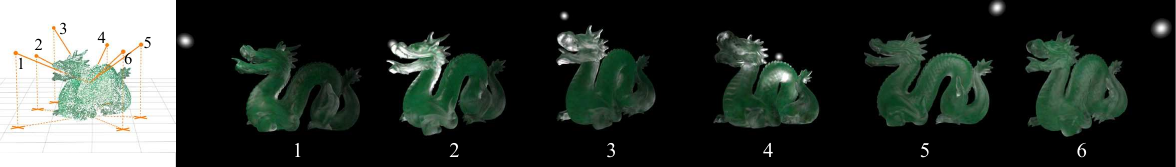}
    \caption{\textbf{Point light relighting}: The leftmost shows the six positions of the point light sources we use to illuminate a translucent \textsc{Dragon}. The renderings of the model illuminated from each position are shown. \textsc{Dragon} becomes brighter as the light gets nearer; please see the supplementary video.}
    \label{fig:moving-point-light}
\end{figure*}

Another strategy we use in the optimization is the late activation of $\Ti$.
$\Ti$ in \cref{eq:out-radiance-sum} is defined as a residual field that only represents the effects unable to be captured by other components, but the optimization could cause $\Ti$ dominating, representing most of the light transport effect.
Therefore, we mitigate this problem by adding $\Ti$ into optimization at the late stage of the optimization (final 30k out of 100k iterations), during which the values of other components already stabilize.

\subsection{Data and Implementation Details}

\textbf{\textit{Synthetic data}}. Our synthetic dataset is generated using 40 different light sources and 48 different cameras, amounting to 1920 images. These lights and cameras are evenly placed on a hemisphere with the subject at the center.
The images in our dataset are comprised of three parts: 1) 40 different OLAT lighting conditions: only one of the light sources is turned on, and each camera generates an image under this OLAT condition, serving as the training set; 2) all light turned on: all the 40 lights are turned on simultaneously. This partition provides a neutral lighting condition and is used to train the original Gaussian Splatting model to obtain a set of Gaussian primitives that serves as the initial values of our optimization loop; 3) 58 novel OLAT lighting conditions with one camera per OLAT on which we evaluate our algorithm.

\begin{figure}[h]
    \centering
    \includegraphics[width=\linewidth]{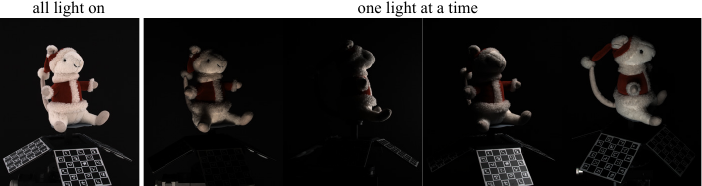}
    \caption{\textbf{Example images} of our capture OLAT dataset.}
    \label{fig:data}
\end{figure}

\noindent\textbf{\textit{Captured data.}}
We also evaluate our method using real-world OLAT data, which is captured with a light stage mounted with 216 global-shutter industrial cameras and 145 programmable LED lights. For each subject, we utilize 88 front-viewing LEDs to capture a total of 19,008 images, and 216 additional images with all lights on for extracting the foreground mask and obtaining the initialization of the training process. Examples of the images can be found in \cref{fig:data}.  
The positions of the cameras and lights are calibrated such that they are in the same coordinate system. \Cref{fig:capture-data-results} shows the intrinsic decomposition and relighting results on the capture dataset.

\noindent\textbf{\textit{Training.}} For simplicity of the training process, we use the all-light-turned-on part of our data to train a Gaussian Splatting model for initializing our training pipeline. We take the Gaussian Splat's opacity, covariance, and positions as the initial values of the same quantities in our model, and the colors as the initial values of $\rho$. 
During training, the number of Gaussians stays constant -- no culling, merging, or splitting. This implies the training quality of our relightable pipeline is limited by the original Gaussian Splatting we start the training with. 
We will leave the exploration of a more systematic training scheme as future work. For training the Gaussian Splat model, The implementation we use is \textsc{Splatfacto} provided in nerfstudio \cite{tancik_nerfstudio_2023}. 

\section{Results}
\label{sec:results}

\begin{figure}[b]
    \centering
    \includegraphics[width=0.9\linewidth]{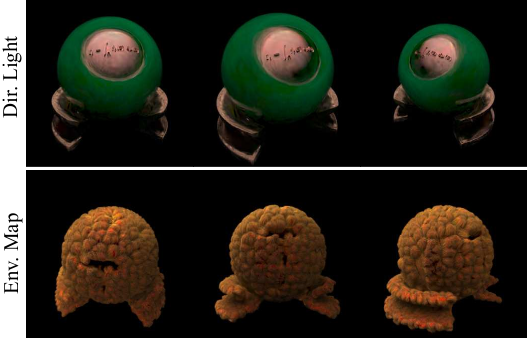}
    \caption{\textbf{Directional light and environment map relighting}. The \textsc{Knob} model is lit with a vertical directional light.  The \textsc{FurBall} model is lit using an environment map. Both are viewed from three different angles.}
    \label{fig:env-light-novel-view}
    \label{fig:distant-light}
\end{figure}

We present experiments on relighting and novel view synthesis on synthetic and capture data. All experiments are conducted on a single NVIDIA A100 GPU. All models are trained with 100k iterations on the OLAT data, taking 3 hours to finish. 

We implemented a CUDA kernel for evaluating the spherical harmonics in \cref{eq:sh-partial-eval,eq:sh-reconstruction}, and the rest of the relighting computation, i.e., \cref{eq:out-radiance-sum} is in PyTorch. With the color computed, the rasterization is done via \textsc{Splatfacto} method in nerfstudio \cite{tancik_nerfstudio_2023}. 

\noindent\textbf{\textit{Model size and runtime.}} 
Each bidirectional Gaussian primitive needs 1,089 optimizable parameters, amounting to \num{4.254} KB memory consumption using 32-bit floating point number. This usually leads to a nearly 200 MB memory cost for a model of around 40,000 primitives. The runtime of our pipeline scales linearly with the number of Gaussian primitives. 
With our hardware, evaluating \cref{eq:out-radiance-sum} takes on average 5--6ms for a model of around 40,000 primitives, dwarfed by the rasterization step which takes around 19.5ms. For the run time and model size of each example used in our paper, please refer to \cref{tab:runtime-size}.

\subsection{Relighting}

Our method applies to different surface and volumetric materials.
As shown in \cref{fig:components}, our method gives plausible relighting and intrinsic lighting decomposition across various types of materials on different objects using a point light. 
The objects we tested include glossy surface-based appearance such as \textsc{MetalBunny} and \textsc{IridescenceBall}; fuzzy material such as \textsc{FurBall} and \textsc{Hairball}; translucent volume such as \textsc{Dragon}.
The method performs especially well on modeling volumetric and fuzzy appearance.
We refer the user to the video in the supplementary material for the results of intrinsic decomposition lit with different point light sources.

\begin{figure*}[ht]
    \centering
    \includegraphics[width=\linewidth]{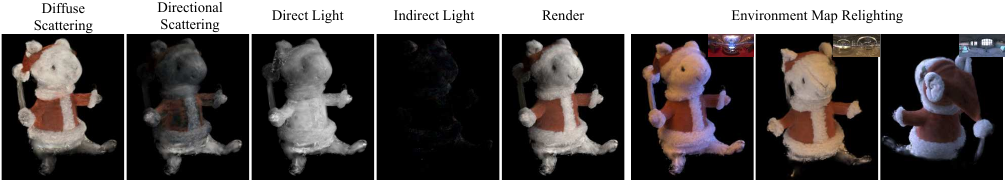}
    \caption{\textbf{Capture data result.} We test our method on the \textsc{Plushy} OLAT data, and perform intrinsic decomposition under a novel point light source, and relight the \textsc{Plushy} with three environment maps, rendering from three distinct viewpoints.}
    \label{fig:capture-data-results}
\end{figure*}

\begin{figure*}[ht]
    \centering
    \includegraphics[width=\linewidth]{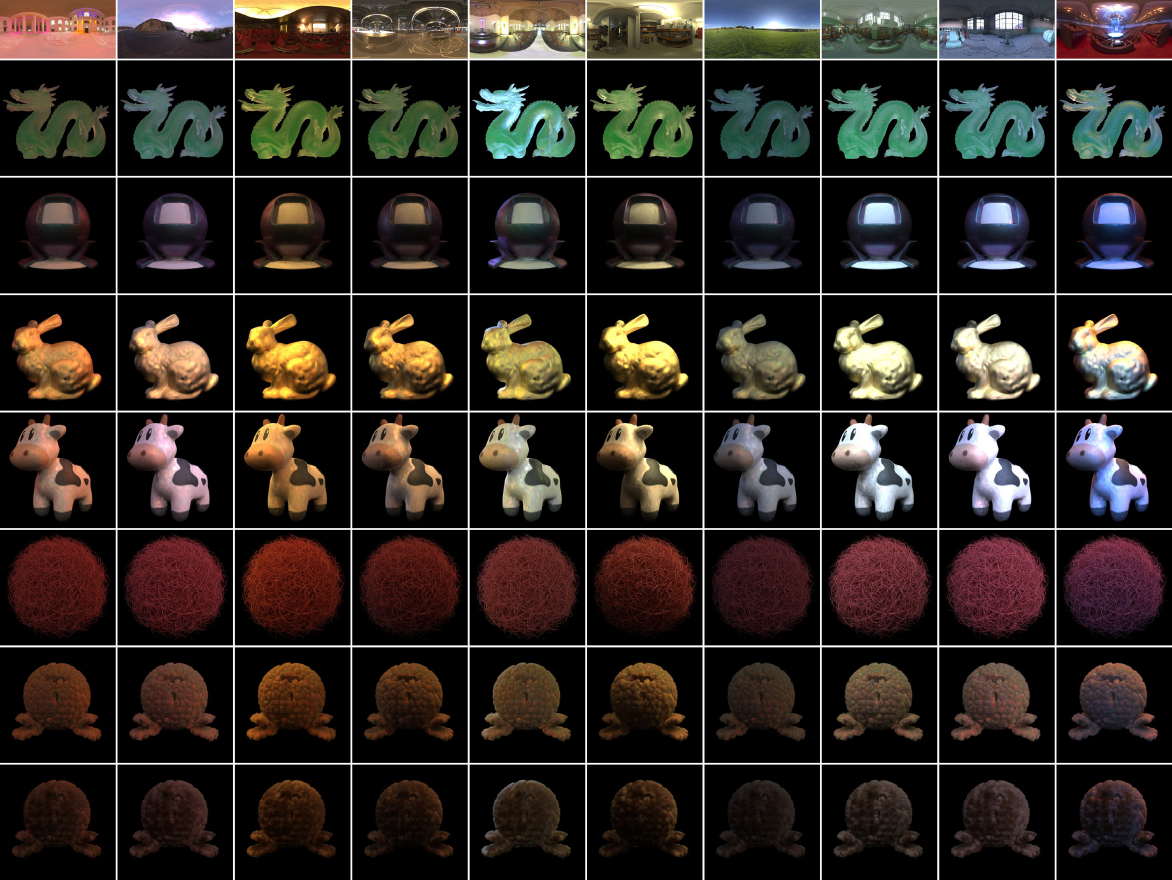}
    \caption{\textbf{Relighting with environment maps.}}
    \label{fig:env-light-gallery}
\end{figure*}

Our method supports relighting an object using point light, directional light, and environment maps. In \cref{fig:moving-point-light}, the translucent \textsc{Dragon} model is lit by a point light, rendered from varying viewpoints. The point light is rotating around with varying distances to the object. The brightness of the model increases as the point light approaches it. In \cref{fig:distant-light}, the \textsc{Knob} with both reflective surface and jade-like material is re-lit with a directional light shone vertically from the top, the reflection of the surface can be observed from the rendering. The \textsc{FurBall} is lit by the environment map from two angles and produces consistent novel view synthesis results. \Cref{fig:capture-data-results} demonstrates both our intrinsic decomposition and environment map relighting on a capture dataset.
\Cref{fig:env-light-gallery} presents relighting results with more environment maps.
Please see the video in the supplementary material for more results.

\begin{figure}[t]
    \centering
    \includegraphics[width=\linewidth]{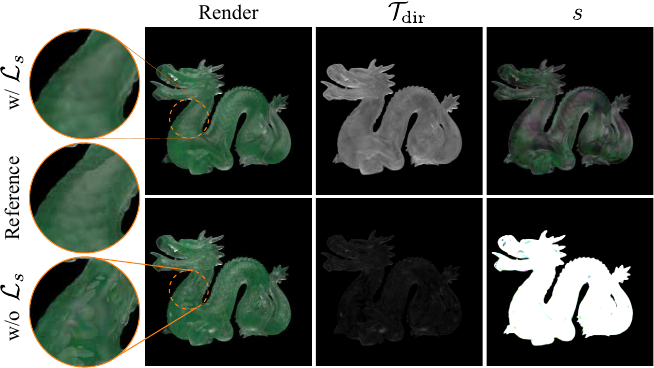}
    \caption{
        \textbf{$\mathcal{L}_\mathrm{s}$ preventing excessive value of $s$}: without constraints, the optimization loop might make $s$ nonphysically large and cause noisy blobs in the rendering. Adding $\mathcal{L}_\mathrm{s}$  alleviates the problem by penalizing overly large $s$. }
    \label{fig:norm-constraint}
\end{figure}

\subsection{Ablation Study}
\begin{figure}[b]
    \centering
    \includegraphics[width=0.95\linewidth]{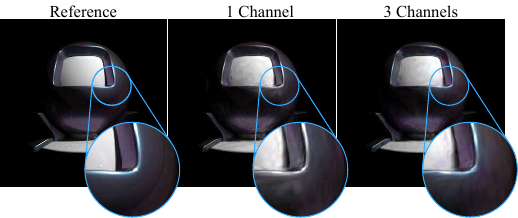}
    \caption{
        \textbf{Number of channels in directional scattering}: using only 1 channel for directional scattering is not able to express blue-tinted reflection from the iridescent material when using 3 channels.
    }
    \label{fig:phase-channels}
\end{figure}

\textbf{The $\mathcal{L}_\mathrm{s}$ loss term} is introduced to constrain the directional scattering $s$ (\cref{eq:norm-constraint}). 
The effect of $\mathcal{L}_\mathrm{s}$ can be seen in \cref{fig:norm-constraint}: without $\mathcal{L}_\mathrm{s}$, the value of $s$ is unchecked and could be very large. This means the amount of directionally scattered light can be many times that of the incoming light, leading to a nonphysical decomposition, and also noise in the directional scattering component and the final rendering.

\noindent\textbf{The number of channels of $s$} affects the specular highlight that our model is able to learn. Using a single channel only generates specular highlights of neutral color, whereas multiple-channel $s$ supports highlights of different tones, such as the blue tint in \textsc{IridescenceBall}  in \cref{fig:phase-channels}.

\begin{table*}[htbp]
    \centering
    \caption{The runtime and model size of our examples, sorted by number of primitives. Our relighting step roughly scales linearly with the number of primitives, providing 40--50 fps of relighting and rendering. Each Gaussian primitive costs 1,089 optimizable parameters, amounting to a per-Gaussian 4.254 KB memory cost using 32-bit floating point numbers.}
    \begin{tabular}{c|ccc|ccc}
                    & \multicolumn{3}{c|}{Time (ms)}    & \multicolumn{3}{c}{Model Size}        \\ 
    Model           & Relight & Rasterize & Total Time & \# Primitives & \# Parameters & Memory (MB) \\ \hline
    \textsc{BunnyMetal} &  \num{1.75}  &\num{19.38}  & \num{21.13} &   \num{16693}  &   \num{18178677} &  \num{69.35}\\ 
    \textsc{Dragon} &  \num{3.71}  & \num{20.96}   &  \num{24.67} &   \num{31252}   &   \num{34033428} & \num{129.83}\\ 
    \textsc{Spot} &  \num{3.83}  &\num{18.52}  & \num{22.35} &   \num{33087}  &   \num{36031743} &  \num{137.45}\\
    \textsc{FurballSpecular} &  \num{4.02}  &\num{19.23}  & \num{23.25} &   \num{34619}  &   \num{37700091} &  \num{143.82}\\ 
    \textsc{IridenscenceBall} & \num{4.02}  & \num{19.20}  & \num{23.21}  &   \num{35035}   &   \num{38153115} & \num{145.54}\\
    \textsc{FurballDiffuse} &  \num{5.11} &\num{19.63} & \num{24.74} & \num{44441}  &\num{48306249} &  \num{184.62}\\ 
    \textsc{Knob} &  \num{6.23}  &\num{19.25}  & \num{25.48} &   \num{53429}  &   \num{58184181} &  \num{221.96}\\ 
    \textsc{HairBall} & \num{15.90}  & \num{19.30}   &  \num{35.20}  &   \num{127787} & \num{139160043} & \num{530.85}\\ 
    \textsc{Plushy} &  \num{19.31}  &\num{17.18}  & \num{36.49} &   \num{153467}  &   \num{167125563} &  \num{637.53}\\ 
    \end{tabular}
    \label{tab:runtime-size}
\end{table*}

\subsection{Comparison} 
\begin{figure}[hb]
    \centering
    \includegraphics[width=\linewidth]{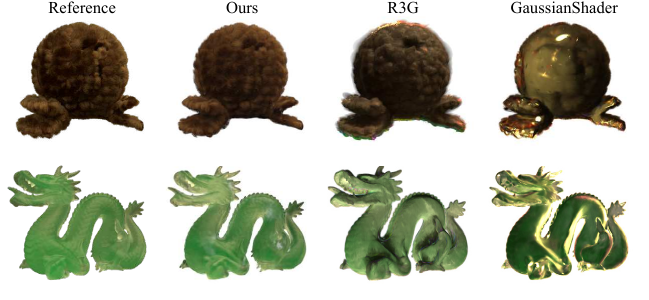}
    \caption{\textbf{Qualitative comparison with R3G \cite{gao_relightable_2023} and GaussianShader \cite{jiang_gaussianshader_2023}}: our model preserves the details of the fur in \textsc{FurBall} and capture subsurface scattering effect in \textsc{Dragon}.}
    \label{fig:comparison}
\end{figure}

We qualitatively compare our method against R3G by Gao et al. \cite{gao_relightable_2023}, and GaussianShader by Jiang et al. \cite{jiang_gaussianshader_2023}. 
It is worth noting that both have different training setups from ours; we leverage OLAT datasets and they require only one environment map lighting during training.
We train both of their methods using a neutral environment map, and relight with another environment map \textsc{GearShop}. Our model restores more volumetric light transport effects like the subsurface scattering in \textsc{Dragon} and preserves the details of the fur in \textsc{FurBall} that can be seen in \cref{fig:comparison}.

\begin{figure*}[ht]
    \centering
    \includegraphics[width=\linewidth]{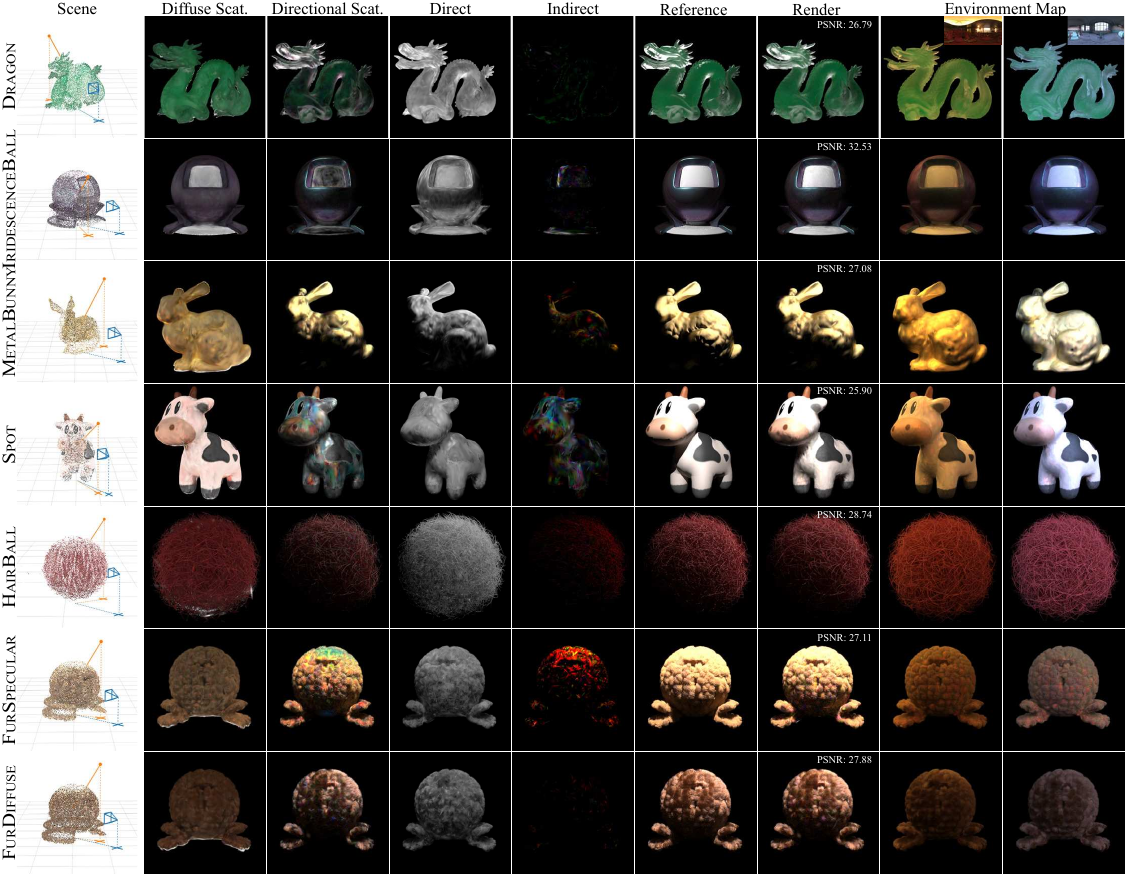}
     \caption{\textbf{Intrinsic decomposition and relighting.} 
     We visualize the intrinsic decomposition components given a novel point light, and relighting under environment maps. From left to right: the scene setups including novel point light positions and camera poses (unseen during training); diffuse scattering; directional scattering; direct transport; indirect transport; the reference images; our renders with PSNR between the references and renders; two renderings under two distinct environment maps. 
     }
    \label{fig:components}
\end{figure*}

\section{Conclusion}
\label{sec:conclusion}

We present a method that enables image synthesis under novel view and lighting conditions of objects represented by Gaussian Splats. Our method decomposes the appearances of the Gaussian primitives into intrinsic components represented by spherical harmonics. 
Our model is able to generate spherical harmonics coefficients for each Gaussian primitive that are compatible with the Gaussian rasterization pipeline, enabling real-time relighting and novel view synthesis performance.
We test our method over OLAT datasets of a variety of materials, showcasing the versatile modeling capability for various appearances.\\

\noindent\textbf{\textit{Limitations and future work}.} 
Our method has some limitations that can be improved in the future.
Even though we propose a few strategies to reduce the ambiguity of different lighting components and increase the stability of the training, 
there are still cases where the intrinsic decomposition does not yield satisfactory results, such as the directional scattering component of \textsc{Spot} in \cref{fig:components}.
To represent different lighting components in our method, we use spherical harmonics, which are an inherently low-frequency representation.
This makes our method struggle to accurately capture some high-frequency light transport effects, such as hard shadow. 
We believe there are multiple promising directions to improve:
inspiration can be taken from prior works from the real-time rendering community, for example, choices of basis functions for representing all-frequency light transport effects \cite{tsai_all-frequency_2006, ng_triple_2004, xu_lightweight_2022}; alternatively, one might condition lighting components on the positions and the opacities of other Gaussian primitives in the scene.
There are also many possibilities to parameterize the scattering functions or incorporate material models to enable more robust modeling of certain light transport effects.

{
    \small
    \bibliographystyle{ieeenat_fullname}
    \bibliography{main}

\begin{thebibliography}{38}
\providecommand{\natexlab}[1]{#1}
\providecommand{\url}[1]{\texttt{#1}}
\expandafter\ifx\csname urlstyle\endcsname\relax
  \providecommand{\doi}[1]{doi: #1}\else
  \providecommand{\doi}{doi: \begingroup \urlstyle{rm}\Url}\fi

\bibitem[Barron et~al.(2022)Barron, Mildenhall, Verbin, Srinivasan, and Hedman]{barron_mip-nerf_2022}
Jonathan~T. Barron, Ben Mildenhall, Dor Verbin, Pratul~P. Srinivasan, and Peter Hedman.
\newblock Mip-{NeRF} 360: {Unbounded} {Anti}-{Aliased} {Neural} {Radiance} {Fields}.
\newblock In \emph{2022 {IEEE}/{CVF} {Conference} on {Computer} {Vision} and {Pattern} {Recognition} ({CVPR})}, pages 5460--5469, New Orleans, LA, USA, 2022. IEEE.

\bibitem[Cai et~al.(2024)Cai, Qiu, Li, and Ren]{cai_neuralto_2024}
Yuxiang Cai, Jiaxiong Qiu, Zhong Li, and Bo Ren.
\newblock {NeuralTO}: {Neural} {Reconstruction} and {View} {Synthesis} of {Translucent} {Objects}.
\newblock \emph{ACM Trans. Graph.}, 43\penalty0 (4):\penalty0 50:1--50:14, 2024.

\bibitem[Condor et~al.(2024)Condor, Speierer, Bode, Bozic, Green, Didyk, and Jarabo]{condor_volumetric_2024}
Jorge Condor, Sebastien Speierer, Lukas Bode, Aljaz Bozic, Simon Green, Piotr Didyk, and Adrian Jarabo.
\newblock Volumetric {Primitives} for {Modeling} and {Rendering} {Scattering} and {Emissive} {Media}, 2024.
\newblock arXiv:2405.15425 [cs].

\bibitem[Gao et~al.(2023)Gao, Gu, Lin, Zhu, Cao, Zhang, and Yao]{gao_relightable_2023}
Jian Gao, Chun Gu, Youtian Lin, Hao Zhu, Xun Cao, Li Zhang, and Yao Yao.
\newblock Relightable {3D} {Gaussian}: {Real}-time {Point} {Cloud} {Relighting} with {BRDF} {Decomposition} and {Ray} {Tracing}, 2023.
\newblock arXiv:2311.16043 [cs].

\bibitem[Ge et~al.(2023)Ge, Hu, Zhao, Liu, and Chen]{ge_ref-neus_2023}
Wenhang Ge, Tao Hu, Haoyu Zhao, Shu Liu, and Ying-Cong Chen.
\newblock Ref-{NeuS}: {Ambiguity}-{Reduced} {Neural} {Implicit} {Surface} {Learning} for {Multi}-{View} {Reconstruction} with {Reflection}.
\newblock pages 4251--4260, 2023.

\bibitem[Guédon and Lepetit(2023)]{guedon_sugar_2023}
Antoine Guédon and Vincent Lepetit.
\newblock {SuGaR}: {Surface}-{Aligned} {Gaussian} {Splatting} for {Efficient} {3D} {Mesh} {Reconstruction} and {High}-{Quality} {Mesh} {Rendering}, 2023.
\newblock arXiv:2311.12775 [cs].

\bibitem[He et~al.(2024)He, Wang, and Yang]{he_gs-phong_2024}
Yumeng He, Yunbo Wang, and Xiaokang Yang.
\newblock {GS}-{Phong}: {Meta}-{Learned} {3D} {Gaussians} for {Relightable} {Novel} {View} {Synthesis}, 2024.
\newblock arXiv:2405.20791 [cs].

\bibitem[Heitz et~al.(2015)Heitz, Dupuy, Crassin, and Dachsbacher]{heitz_sggx_2015}
Eric Heitz, Jonathan Dupuy, Cyril Crassin, and Carsten Dachsbacher.
\newblock The {SGGX} microflake distribution.
\newblock \emph{ACM Transactions on Graphics}, 34\penalty0 (4):\penalty0 48:1--48:11, 2015.

\bibitem[Huang et~al.(2024)Huang, Yu, Chen, Geiger, and Gao]{huang_2d_2024}
Binbin Huang, Zehao Yu, Anpei Chen, Andreas Geiger, and Shenghua Gao.
\newblock {2D} {Gaussian} {Splatting} for {Geometrically} {Accurate} {Radiance} {Fields}, 2024.
\newblock arXiv:2403.17888 [cs].

\bibitem[Jeong et~al.(2024)Jeong, Koo, Zhang, and Kim]{jeong_esr-nerf_2024}
Jinseo Jeong, Junseo Koo, Qimeng Zhang, and Gunhee Kim.
\newblock {ESR}-{NeRF}: {Emissive} {Source} {Reconstruction} {Using} {LDR} {Multi}-view {Images}, 2024.
\newblock arXiv:2404.15707 [cs].

\bibitem[Jiang et~al.(2023)Jiang, Tu, Liu, Gao, Long, Wang, and Ma]{jiang_gaussianshader_2023}
Yingwenqi Jiang, Jiadong Tu, Yuan Liu, Xifeng Gao, Xiaoxiao Long, Wenping Wang, and Yuexin Ma.
\newblock {GaussianShader}: {3D} {Gaussian} {Splatting} with {Shading} {Functions} for {Reflective} {Surfaces}, 2023.
\newblock arXiv:2311.17977 [cs].

\bibitem[Kerbl et~al.(2023)Kerbl, Kopanas, Leimkuehler, and Drettakis]{kerbl_3d_2023}
Bernhard Kerbl, Georgios Kopanas, Thomas Leimkuehler, and George Drettakis.
\newblock {3D} {Gaussian} {Splatting} for {Real}-{Time} {Radiance} {Field} {Rendering}.
\newblock \emph{ACM Transactions on Graphics}, 42\penalty0 (4):\penalty0 139:1--139:14, 2023.

\bibitem[Lehtinen(2007)]{lehtinen_framework_2007}
Jaakko Lehtinen.
\newblock A framework for precomputed and captured light transport.
\newblock \emph{ACM Trans. Graph.}, 26\penalty0 (4):\penalty0 13--es, 2007.

\bibitem[Liu et~al.(2023)Liu, Wang, Lin, Long, Wang, Liu, Komura, and Wang]{liu_nero_2023}
Yuan Liu, Peng Wang, Cheng Lin, Xiaoxiao Long, Jiepeng Wang, Lingjie Liu, Taku Komura, and Wenping Wang.
\newblock {NeRO}: {Neural} {Geometry} and {BRDF} {Reconstruction} of {Reflective} {Objects} from {Multiview} {Images}.
\newblock \emph{ACM Transactions on Graphics}, 42\penalty0 (4):\penalty0 1--22, 2023.

\bibitem[Lyu et~al.(2022)Lyu, Tewari, Leimkuehler, Habermann, and Theobalt]{lyu_neural_2022}
Linjie Lyu, Ayush Tewari, Thomas Leimkuehler, Marc Habermann, and Christian Theobalt.
\newblock Neural {Radiance} {Transfer} {Fields} for {Relightable} {Novel}-view {Synthesis} with {Global} {Illumination}, 2022.
\newblock arXiv:2207.13607 [cs].

\bibitem[Mildenhall et~al.(2020)Mildenhall, Srinivasan, Tancik, Barron, Ramamoorthi, and Ng]{mildenhall_nerf_2020}
Ben Mildenhall, Pratul~P. Srinivasan, Matthew Tancik, Jonathan~T. Barron, Ravi Ramamoorthi, and Ren Ng.
\newblock {NeRF}: {Representing} {Scenes} as {Neural} {Radiance} {Fields} for {View} {Synthesis}.
\newblock In \emph{{ECCV}}, 2020.

\bibitem[Ng et~al.(2004)Ng, Ramamoorthi, and Hanrahan]{ng_triple_2004}
Ren Ng, Ravi Ramamoorthi, and Pat Hanrahan.
\newblock Triple product wavelet integrals for all-frequency relighting.
\newblock \emph{ACM Transactions on Graphics}, 23\penalty0 (3):\penalty0 477--487, 2004.

\bibitem[Saito et~al.(2023)Saito, Schwartz, Simon, Li, and Nam]{saito_relightable_2023}
Shunsuke Saito, Gabriel Schwartz, Tomas Simon, Junxuan Li, and Giljoo Nam.
\newblock Relightable {Gaussian} {Codec} {Avatars}, 2023.
\newblock arXiv:2312.03704 [cs].

\bibitem[Sarkar et~al.(2023)Sarkar, Bühler, Li, Wang, Vicini, Riviere, Zhang, Orts-Escolano, Gotardo, Beeler, and Meka]{sarkar_litnerf_2023}
Kripasindhu Sarkar, Marcel~C. Bühler, Gengyan Li, Daoye Wang, Delio Vicini, Jérémy Riviere, Yinda Zhang, Sergio Orts-Escolano, Paulo Gotardo, Thabo Beeler, and Abhimitra Meka.
\newblock {LitNeRF}: {Intrinsic} {Radiance} {Decomposition} for {High}-{Quality} {View} {Synthesis} and {Relighting} of {Faces}.
\newblock In \emph{{SIGGRAPH} {Asia} 2023 {Conference} {Papers}}, pages 1--11, Sydney NSW Australia, 2023. ACM.

\bibitem[Sloan et~al.(2002)Sloan, Kautz, and Snyder]{sloan_precomputed_2002}
Peter-Pike Sloan, Jan Kautz, and John Snyder.
\newblock Precomputed radiance transfer for real-time rendering in dynamic, low-frequency lighting environments.
\newblock \emph{ACM Transactions on Graphics}, 21\penalty0 (3):\penalty0 527--536, 2002.

\bibitem[Srinivasan et~al.(2020)Srinivasan, Deng, Zhang, Tancik, Mildenhall, and Barron]{srinivasan_nerv_2020}
Pratul~P. Srinivasan, Boyang Deng, Xiuming Zhang, Matthew Tancik, Ben Mildenhall, and Jonathan~T. Barron.
\newblock {NeRV}: {Neural} {Reflectance} and {Visibility} {Fields} for {Relighting} and {View} {Synthesis}, 2020.
\newblock arXiv:2012.03927 [cs].

\bibitem[Tancik et~al.(2023)Tancik, Weber, Ng, Li, Yi, Wang, Kristoffersen, Austin, Salahi, Ahuja, Mcallister, Kerr, and Kanazawa]{tancik_nerfstudio_2023}
Matthew Tancik, Ethan Weber, Evonne Ng, Ruilong Li, Brent Yi, Terrance Wang, Alexander Kristoffersen, Jake Austin, Kamyar Salahi, Abhik Ahuja, David Mcallister, Justin Kerr, and Angjoo Kanazawa.
\newblock Nerfstudio: {A} {Modular} {Framework} for {Neural} {Radiance} {Field} {Development}.
\newblock In \emph{{ACM} {SIGGRAPH} 2023 {Conference} {Proceedings}}, pages 1--12, New York, NY, USA, 2023. Association for Computing Machinery.

\bibitem[Toschi et~al.(2023)Toschi, De~Matteo, Spezialetti, De~Gregorio, Di~Stefano, and Salti]{toschi_relight_2023}
Marco Toschi, Riccardo De~Matteo, Riccardo Spezialetti, Daniele De~Gregorio, Luigi Di~Stefano, and Samuele Salti.
\newblock {ReLight} {My} {NeRF}: {A} {Dataset} for {Novel} {View} {Synthesis} and {Relighting} of {Real} {World} {Objects}.
\newblock In \emph{2023 {IEEE}/{CVF} {Conference} on {Computer} {Vision} and {Pattern} {Recognition} ({CVPR})}, pages 20762--20772, Vancouver, BC, Canada, 2023. IEEE.

\bibitem[Tsai and Shih(2006)]{tsai_all-frequency_2006}
Yu-Ting Tsai and Zen-Chung Shih.
\newblock All-frequency precomputed radiance transfer using spherical radial basis functions and clustered tensor approximation.
\newblock \emph{ACM Transactions on Graphics}, 25\penalty0 (3):\penalty0 967--976, 2006.

\bibitem[Verbin et~al.(2021)Verbin, Hedman, Mildenhall, Zickler, Barron, and Srinivasan]{verbin_ref-nerf_2021}
Dor Verbin, Peter Hedman, Ben Mildenhall, Todd Zickler, Jonathan~T. Barron, and Pratul~P. Srinivasan.
\newblock Ref-{NeRF}: {Structured} {View}-{Dependent} {Appearance} for {Neural} {Radiance} {Fields}, 2021.
\newblock arXiv:2112.03907 [cs].

\bibitem[Wang et~al.(2023)Wang, Zhang, and Süsstrunk]{wang_nemto_2023}
Dongqing Wang, Tong Zhang, and Sabine Süsstrunk.
\newblock {NEMTO}: {Neural} {Environment} {Matting} for {Novel} {View} and {Relighting} {Synthesis} of {Transparent} {Objects}.
\newblock In \emph{2023 {IEEE}/{CVF} {International} {Conference} on {Computer} {Vision} ({ICCV})}, pages 317--327, Paris, France, 2023. IEEE.

\bibitem[Wang et~al.(2021)Wang, Liu, Liu, Theobalt, Komura, and Wang]{wang_neus_2021}
Peng Wang, Lingjie Liu, Yuan Liu, Christian Theobalt, Taku Komura, and Wenping Wang.
\newblock {NeuS}: {Learning} {Neural} {Implicit} {Surfaces} by {Volume} {Rendering} for {Multi}-view {Reconstruction}.
\newblock In \emph{Advances in {Neural} {Information} {Processing} {Systems}}, pages 27171--27183. Curran Associates, Inc., 2021.

\bibitem[Xu et~al.(2023)Xu, Zoss, Chandran, Gross, Bradley, and Gotardo]{xu_renerf_2023}
Yingyan Xu, Gaspard Zoss, Prashanth Chandran, Markus Gross, Derek Bradley, and Paulo Gotardo.
\newblock {ReNeRF}: {Relightable} {Neural} {Radiance} {Fields} with {Nearfield} {Lighting}.
\newblock 2023.

\bibitem[Xu et~al.(2022)Xu, Zeng, Wu, Wang, and Yan]{xu_lightweight_2022}
Zilin Xu, Zheng Zeng, Lifan Wu, Lu Wang, and Ling-Qi Yan.
\newblock Lightweight {Neural} {Basis} {Functions} for {All}-{Frequency} {Shading}.
\newblock In \emph{{SIGGRAPH} {Asia} 2022 {Conference} {Papers}}, pages 1--9, Daegu Republic of Korea, 2022. ACM.

\bibitem[Yariv et~al.(2021)Yariv, Gu, Kasten, and Lipman]{yariv_volume_2021}
Lior Yariv, Jiatao Gu, Yoni Kasten, and Yaron Lipman.
\newblock Volume {Rendering} of {Neural} {Implicit} {Surfaces}, 2021.
\newblock arXiv:2106.12052 [cs].

\bibitem[Yariv et~al.(2023)Yariv, Hedman, Reiser, Verbin, Srinivasan, Szeliski, Barron, and Mildenhall]{yariv_bakedsdf_2023}
Lior Yariv, Peter Hedman, Christian Reiser, Dor Verbin, Pratul~P. Srinivasan, Richard Szeliski, Jonathan~T. Barron, and Ben Mildenhall.
\newblock {BakedSDF}: {Meshing} {Neural} {SDFs} for {Real}-{Time} {View} {Synthesis}.
\newblock In \emph{{ACM} {SIGGRAPH} 2023 {Conference} {Proceedings}}, pages 1--9, New York, NY, USA, 2023. Association for Computing Machinery.

\bibitem[Yu et~al.(2023)Yu, Guo, Fathi, Chang, Chan, Gao, Funkhouser, and Wu]{yu_learning_2023}
Hong-Xing Yu, Michelle Guo, Alireza Fathi, Yen-Yu Chang, Eric~Ryan Chan, Ruohan Gao, Thomas Funkhouser, and Jiajun Wu.
\newblock Learning {Object}-{Centric} {Neural} {Scattering} {Functions} for {Free}-{Viewpoint} {Relighting} and {Scene} {Composition}, 2023.
\newblock arXiv:2303.06138 [cs].

\bibitem[Zeng et~al.(2023)Zeng, Chen, Dong, Peers, Wu, and Tong]{zeng_relighting_2023}
Chong Zeng, Guojun Chen, Yue Dong, Pieter Peers, Hongzhi Wu, and Xin Tong.
\newblock Relighting {Neural} {Radiance} {Fields} with {Shadow} and {Highlight} {Hints}.
\newblock In \emph{{ACM} {SIGGRAPH} 2023 {Conference} {Proceedings}}, pages 1--11, New York, NY, USA, 2023. Association for Computing Machinery.

\bibitem[Zhang et~al.(2021)Zhang, Srinivasan, Deng, Debevec, Freeman, and Barron]{zhang_nerfactor_2021}
Xiuming Zhang, Pratul~P. Srinivasan, Boyang Deng, Paul Debevec, William~T. Freeman, and Jonathan~T. Barron.
\newblock {NeRFactor}: {Neural} {Factorization} of {Shape} and {Reflectance} {Under} an {Unknown} {Illumination}.
\newblock \emph{ACM Transactions on Graphics}, 40\penalty0 (6):\penalty0 1--18, 2021.
\newblock arXiv:2106.01970 [cs].

\bibitem[Zhang et~al.(2023)Zhang, Xu, Yu, Ye, Jing, Wang, Yu, and Yang]{zhang_nemf_2023}
Youjia Zhang, Teng Xu, Junqing Yu, Yuteng Ye, Yanqing Jing, Junle Wang, Jingyi Yu, and Wei Yang.
\newblock {NeMF}: {Inverse} {Volume} {Rendering} with {Neural} {Microflake} {Field}.
\newblock In \emph{2023 {IEEE}/{CVF} {International} {Conference} on {Computer} {Vision} ({ICCV})}, pages 22862--22872, Paris, France, 2023. IEEE.

\bibitem[Zheng et~al.(2021)Zheng, Singh, and Seidel]{zheng_neural_2021}
Quan Zheng, Gurprit Singh, and Hans-peter Seidel.
\newblock Neural {Relightable} {Participating} {Media} {Rendering}.
\newblock In \emph{Advances in {Neural} {Information} {Processing} {Systems}}, pages 15203--15215. Curran Associates, Inc., 2021.

\bibitem[Zhou et~al.(2024)Zhou, Wu, and Yan]{zhou_unified_2024}
Yang Zhou, Songyin Wu, and Ling-Qi Yan.
\newblock Unified {Gaussian} {Primitives} for {Scene} {Representation} and {Rendering}, 2024.
\newblock arXiv:2406.09733 [cs].

\bibitem[Zhu et~al.(2023)Zhu, Saito, Bozic, Aliaga, Darrell, and Lassner]{zhu_neural_2023}
Shizhan Zhu, Shunsuke Saito, Aljaz Bozic, Carlos Aliaga, Trevor Darrell, and Christoph Lassner.
\newblock Neural {Relighting} with {Subsurface} {Scattering} by {Learning} the {Radiance} {Transfer} {Gradient}, 2023.
\newblock arXiv:2306.09322 [cs].

\end{thebibliography}
}
\appendix
\clearpage
\setcounter{page}{1}
\maketitlesupplementary

\section{Reciprocity of $s$}
\label{sec:reciprocity}
We represent $s$ using bidirectional spherical harmonics as in \cref{eq:double-sh}. We intend to prove if $c_{ij} = c_{ji}$, then $s$ is reciprocal, mathematically, $s(\win, \wout) = s(\wout, \win)$.
We rewrite \cref{eq:double-sh} by merging summation terms of index $(i, j)$ and $(j, i)$, having
\begin{equation*}
\begin{split}
    s(\win, \wout) =& \sum_{i = 1}^{n} \sum_{j = i + 1}^{n} \left[ c_{ij} y_i(\win) y_j(\wout) + c_{ji} y_j(\win) y_i(\wout) \right] \\
    +& \sum_{i = 1}^{n} c_{ii} y_i(\win) y_i(\wout)
\end{split}
\end{equation*}

The second summation is the same for $s(\win, \wout)$ and $s(\wout, \win)$. To equate the first summation, we need to have $c_{ij} y_i(\win) y_j(\wout) + c_{ji} y_j(\win) y_i(\wout)$ equal to $c_{ji} y_j(\wout) y_i(\win) + c_{ij} y_j(\wout) y_i(\win)$. Then $c_{ij} = c_{ji}$ gives us the following:
\begin{equation*}
\begin{split}
    & c_{ij} y_i(\win) y_j(\wout) + c_{ji} y_j(\win) y_i(\wout) \\
  = & c_{ji} y_i(\win) y_j(\wout) + c_{ij} y_i(\win) y_j(\wout)\\
  = & c_{ji} y_j(\wout) y_i(\win) + c_{ij} y_j(\wout) y_i(\win).
\end{split}
\end{equation*}
$\hfill\square$

\end{document}